\def\paperlanguage{} 

\newcommand{\switchlanguage}[2]{%
  \ifx\paperlanguage\empty%
  #1%
  \else%
  #2%
  \fi%
}

\pdfoutput=1
\PassOptionsToPackage{pdftex}{graphicx}
\documentclass{tADR2e}

\usepackage{subcaption}
\usepackage{multirow}
\makeatletter
\providecommand{\@bibsetup}[1]{}
\providecommand{\ltx@label}[1]{%
  \@bsphack
  \protected@write\@auxout{}%
    {\string\newlabel{#1}{{\@currentlabel}{\thepage}}}%
  \@esphack
}
\renewcommand{\label}[1]{\ltx@label{#1}}
\newcommand{\AR@trim}[1]{\zap@space#1 \@empty}
\newcommand{\AR@citenum}[1]{%
  \edef\AR@key{\AR@trim{#1}}%
  \@ifundefined{b@\AR@key}{?}{\csname b@\AR@key\endcsname}%
}
\newcommand{\ARcite}[1]{%
  \begingroup
  \def\AR@sep{}%
  \@for\AR@k:=#1\do{%
    \AR@sep\AR@citenum{\AR@k}%
    \def\AR@sep{, }%
  }%
  \endgroup
}
\renewcommand{\cite}[1]{[\ARcite{#1}]}
\makeatother

\begingroup
\begin{document}

\jvol{39} \jnum{12} \jyear{2025} \jmonth{}

\switchlanguage%
  {
  \title{Dexterous Grasp Data Augmentation based on Grasp Synthesis with Fingertip Workspace Cloud and Contact-Aware Sampling}
  }
  {
  \title{Paper Title}
  }

\author{Liqi Wu$^{a}$$^{\ast}$\thanks{$^\ast$Corresponding author. Email: wu@jsk.imi.i.u-tokyo.ac.jp \vspace{6pt}}, Haoyu Jia$^{a}$, Kento Kawaharazuka$^{a}$, Hirokazu Ishida$^{a}$, Kei Okada$^{a}$ \\\vspace{6pt}  $^{a}${\em{The Department of Mechano-Informatics, Graduate School of Information Science and Technology, The University of Tokyo, Bunkyo-ku, Tokyo, Japan}};
\\\vspace{6pt}\received{Published in 2025} }

\maketitle
{\footnotesize
\noindent\textit{This is a preprint of an article whose final and definitive form has been published in ADVANCED ROBOTICS 2025, copyright Taylor \& Francis and Robotics Society of Japan, is available online at: \url{https://doi.org/10.1080/01691864.2025.2524553}.}\par}
\vspace{0.8\baselineskip}

\begin{abstract}
\switchlanguage%
{
Robotic grasping is a fundamental yet crucial component of robotic applications, as effective grasping often serves as the starting point for various tasks. With the rapid advancement of neural networks, data-driven approaches for robotic grasping have become mainstream. However, efficiently generating grasp datasets for training remains a bottleneck. This is compounded by the diverse structures of robotic hands, making the design of generalizable grasp generation methods even more complex. In this work, we propose a teleoperation-based framework to collect a small set of grasp pose demonstrations, which are augmented using FSG--a Fingertip-contact-aware Sampling-based Grasp generator. Based on the demonstrated grasp poses, we propose AutoWS, which automatically generates structured workspace clouds of robotic fingertips, embedding the hand structure information directly into the clouds to eliminate the need for inverse kinematics calculations. Experiments on grasping the YCB objects show that our method significantly outperforms existing approaches in both speed and valid pose generation rate. Our framework enables real-time grasp generation for hands with arbitrary structures and produces human-like grasps when combined with demonstrations, providing an efficient and robust data augmentation tool for data-driven grasp training.
}
{
Within 200 words.
}
\end{abstract}

\begin{keywords}
Grasp Data Augmentation, Grasp Synthesis, Fingertip-Contact-Aware Sampling, Workspace cloud
\end{keywords}

\section{Introduction}
\begin{figure}
  \centering
  \includegraphics[width=\columnwidth]{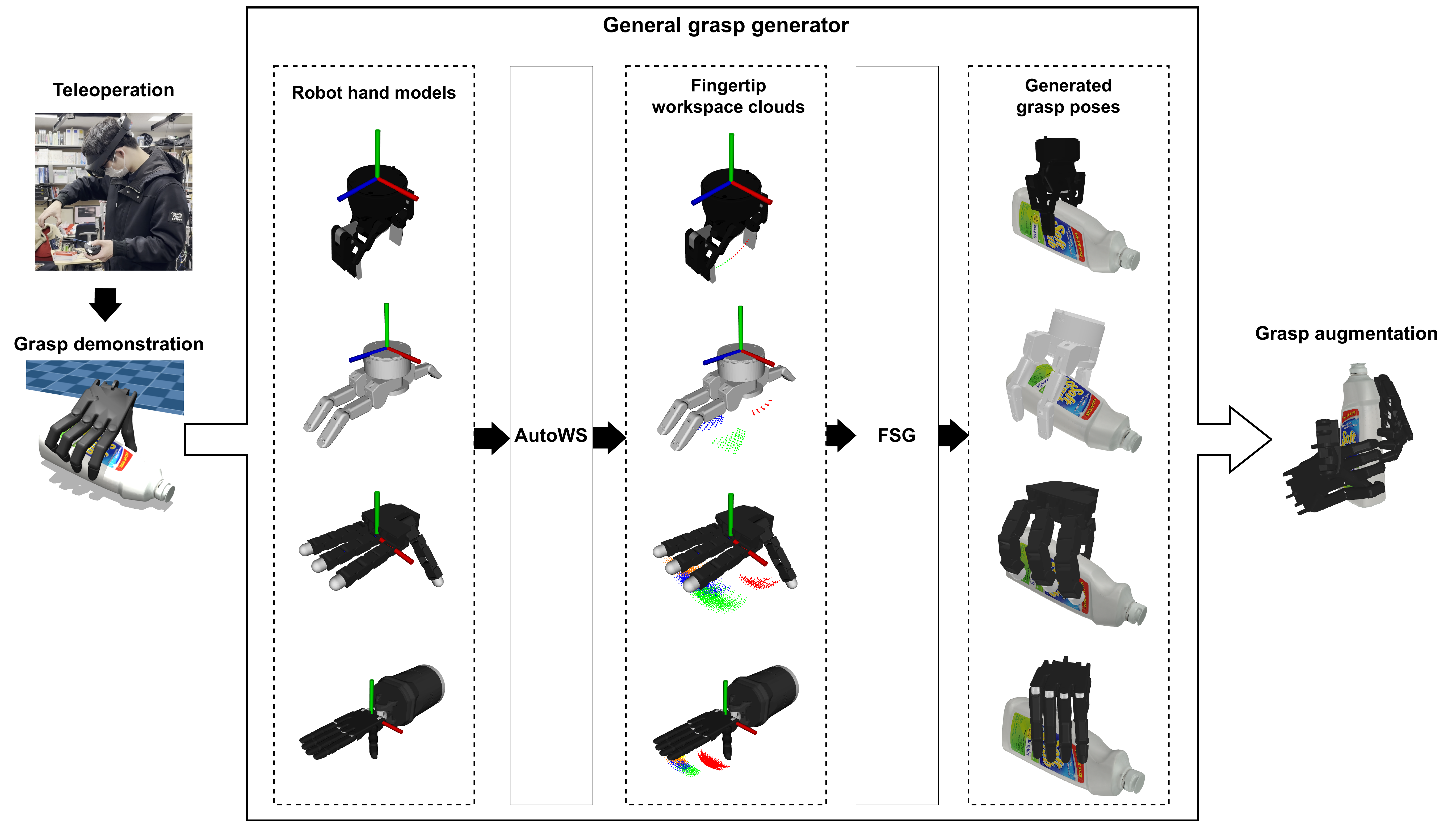}
  \caption{System architecture for dexterous grasp data augmentation. The general grasp generator consists of two components: AutoWS and FSG. Given a robot hand model, AutoWS automatically generates the fingertip workspace clouds. Based on the generated workspace clouds and the shape information of the target object, FSG rapidly generates valid grasp poses through fingertip-contact-aware random sampling. When a human-demonstrated grasp pose is provided, AutoWS generates the workspace clouds corresponding to the demonstrated pose, and FSG generates grasp data with finger joint angles similar to the demonstration but with different grasp locations, enabling grasp data augmentation.}
  \label{fig:whole}
\end{figure}

\switchlanguage%
  {
Robotic grasp synthesis, one of the cornerstones of robotics research, has been a focal area for decades. Much of the previous research \cite{anti1,fin21,fin22,fin23} has focused on grasping strategies for parallel grippers due to their simplicity and stability. However, these approaches often struggle to generalize to multi-fingered dexterous hands with varying structures. Some studies \cite{approach1,exhau3,primitive} focus on multi-fingered hand grasping, overcoming the challenges posed by the high degrees of freedom (DOF) in these hands through the training of neural networks. However, training these networks requires extensive datasets, and the data collection process is often time-consuming and labor-intensive.

The continuous innovation in robotic hand designs underscores the need for a universally applicable and efficient grasp dataset generator rather than hand-specific algorithms. To address the diversity of robotic hand structures, some model-based grasp synthesis algorithms have been proposed. Optimization-based approaches \cite{fast,dex} excel at generating stable grasp poses but often suffer from long computation times. Moreover, these methods typically minimize the distance between the hand and the object, which can lead to challenges in scenarios, such as when grasping small objects from a table, where direct enclosure of the object is hindered by collisions with the surface. Sampling-based methods \cite{billion,pds} offer more flexibility in pose generation but struggle with the vast solution space, resulting in high computational costs and inefficiency for large-scale data generation.

AR and VR technologies have strengthened robotic teleoperation, enabling the capture of natural, human-like poses and motions \cite{tele}. Yet, manually collecting large datasets remains a time- and labor-intensive process. In computer vision, data augmentation \cite{aug} is often used to overcome the challenge of limited data. Could a similar pipeline be leveraged to augment grasp data with limited initial samples?

To address this, we propose AutoWS, a novel approach that utilizes demonstrations of grasp poses and robotic hand structures to automatically generate corresponding fingertip workspace clouds. Building upon this, we introduce FSG, a Fingertip-contact-aware Sampling-based Grasp generator, to generate grasp poses within seconds, enabling the rapid generation of human-like grasp data for augmentation. The system architecture is illustrated in Fig.\ref{fig:whole}.

The main contributions of this work are as follows:
\begin{itemize}
  \item We propose AutoWS, which embeds finger structural information into fingertip workspace clouds derived from robotic hand models and discretizes the solution space for efficient sampling. When combined with teleoperated human data, AutoWS further narrows the solution space, which enables faster generation of human-like grasp poses.
  \item We introduce FSG, which significantly reduces the computation time for sampling-based grasp pose generation while improving the efficiency of valid pose discovery. This approach is applicable to robotic hands of arbitrary structures.
  \item We demonstrate the versatility of our approach through validation on four robotic hands with varying structures and finger counts. Additionally, we showcase the efficacy and speed of our approach for data augmentation using teleoperated data on a five-finger dexterous hand. Comparisons with existing methods highlight the advantages of our approach in terms of efficiency in valid pose generation, while maintaining grasp quality.
\end{itemize}
}
{
Within 6000 words.
}

\section{Related Work}\label{sec:related}
\switchlanguage%
{

Methods for robotic grasp synthesis can be categorized into two main branches: model-based and data-driven methods. Data-driven methods typically require a large amount of robot grasp pose data to train neural networks. To collect this data, model-based methods are often employed. This work focuses on the efficient generation and augmentation of grasp data. Therefore, in this section, we primarily review previous studies on model-based grasp pose generation methods and the data collection aspects of data-driven methods.

\subsection{Sampling-based Grasp Synthesis}
Sampling-based methods \cite{billion} can generally be divided into two categories: contact point-based sampling and approach direction-based sampling.

In the special case of contact point-based sampling, antipodal point sampling was first proposed in \cite{antipodal}, with numerous improvements emerging subsequently \cite{anti2,anti3,anti4}. However, this method is limited to parallel grippers and is difficult to apply directly to multi-finger hands. To address this limitation, some studies \cite{exhau1,exhau2,exhau3} employed exhaustive sampling to generate a series of contact point combinations. These combinations were then filtered based on grasp stability and mechanical structure reachability to generate grasp poses. Despite its effectiveness, exhaustive sampling remains highly time-consuming. To improve efficiency, \cite{pds} utilized Poisson Disk Sampling (PDS) to select 256 points from the object surface, identifying grasp poses that satisfy both force closure and reachability. However, even with a reduced number of sampled points, the number of combinations remain vast. When extended to four-finger or five-finger hands, the computational cost becomes excessive, resulting in low generation efficiency. Consequently, prior research using contact point sampling often limits the number of fingers to three.

Approach direction-based sampling helps mitigate the impact of the number of fingers on generation time. The method proposed in \cite{approach} samples points on the object surface and derives the approach direction from the surface normal at each point. The hand then continuously attempts to grasp the object as it approaches. The stability of each grasp attempt is evaluated, and effective grasp poses are retained during this process. Subsequent studies \cite{approach2,approach3,approach4} have utilized similar methods for data generation. However, determining which points on the hand to approach and adjusting the hand pitch angle require complex tuning. To enhance efficiency further, recent work from \cite{qd} integrates approach direction-based sampling methods with the Quality Diversity algorithm to accelerate grasp pose generation.

\subsection{Optimization-based Grasp Synthesis}

EigenGrasp \cite{eigen} is a classic optimization-based algorithm for grasp pose generation. Renowned for its versatility, it has been widely adopted in previous studies to generate grasp datasets \cite{eigen1,eigen2,eigen3,eigen4}. EigenGrasp reduces the dimensionality of the solution space by leveraging predefined eigen grasp components to model grasp poses. It then employs the simulated annealing algorithm to identify higher-quality grasp poses. However, despite dimensionality reduction, the algorithm remains computationally intensive, and even with numerous optimizations, the quality of the generated poses often stays mediocre.

The problem of grasp pose generation is reformulated in \cite{fast} as a gradient-based optimization of grasp stability under joint angle constraints. Similarly, \cite{dex} defined an energy function that incorporates force closure, finger penetration, and out-of-limit joint angle criteria to optimize the grasp poses for data generation. The resulting datasets has since been used in subsequent data-driven training studies \cite{dex1,dex2}. Additionally, \cite{contact,contact1,contact2} transformed the grasp synthesis problem into optimizing hand-object contact states based on prior contact maps, which generates human-like grasps but introduces the additional requirement for contact map datasets.

}
{
}

\begin{figure*}
  \centering
  \includegraphics[width=\textwidth]{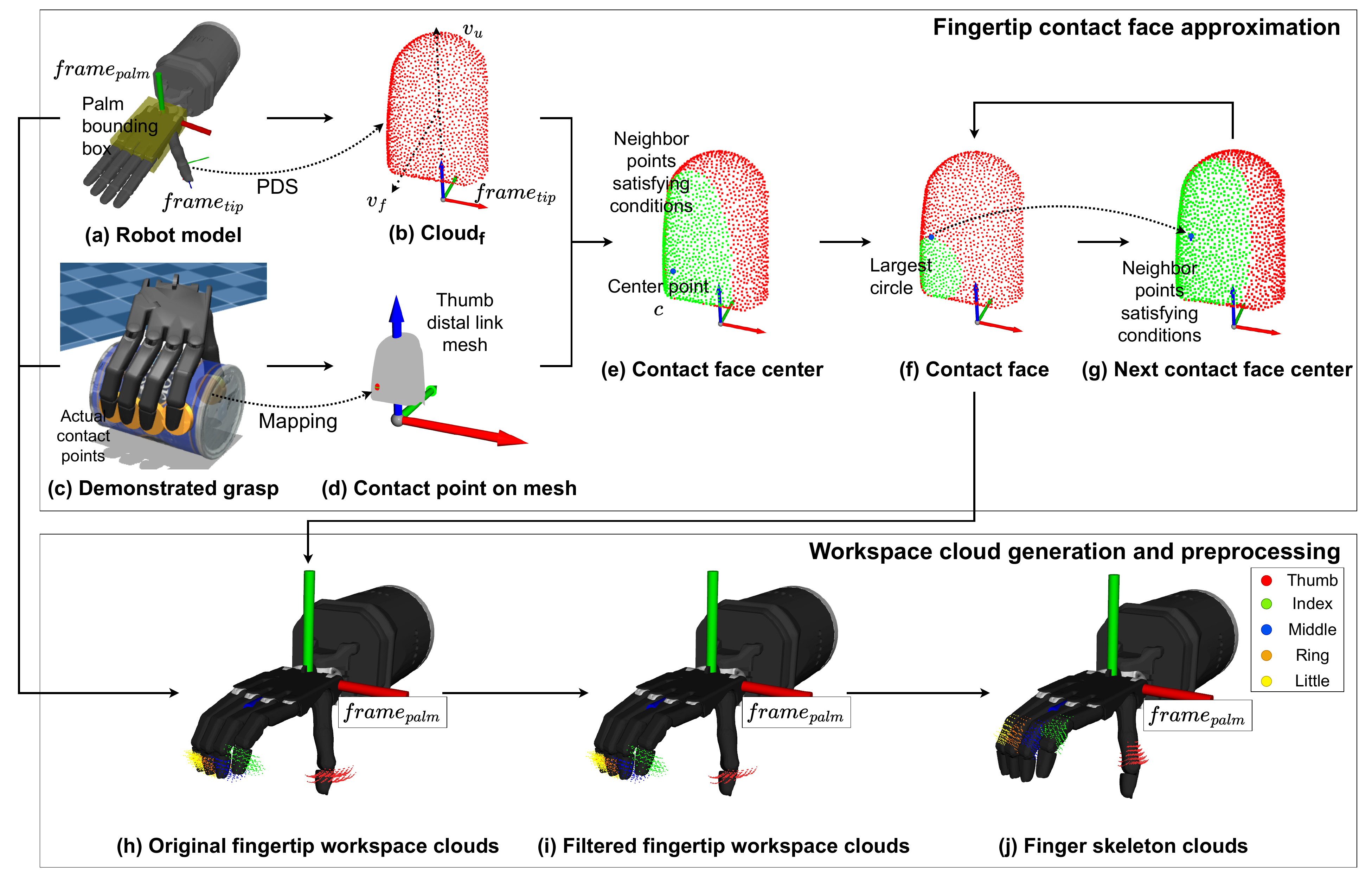}
  \caption{Flowchart of AutoWS. (a) Starting with a robot hand model, (b) the distal link cloud is sampled using PDS and preprocessed based on two vectors, \(v_{u}\)(upward) and \(v_{f}\)(forward), in \(frame_{tip}\). (c-d) Combined with a demonstrated grasp, (e-g) contact faces are approximated. Using the hand structure and joint space information, (h-i) fingertip workspace clouds in \(frame_{palm}\) are generated and filtered, along with (j) corresponding finger skeleton clouds. Each point in the workspace cloud represents a potential contact point for the fingertip, while the line segments between point pairs in the skeleton clouds represents the finger links, excluding fingertips, at specific joint angles.}
  \label{fig:autows}
\end{figure*}

\section{Automatic Generation of Fingertip Workspace Clouds} \label{sec:ws_gen}
\switchlanguage%
{

In this work, we focus on precision grasp using fingertips. The three key components of a precision grasp pose are: the relative pose between the object and the palm, the contact points of the fingertips on the object, and the corresponding joint angles of the fingers at these contact points. In existing approaches based on fingertip contact point sampling \cite{exhau1,pds}, the sampling of contact points and computation of the inverse kinematics (IK) for finger reachability are often performed separately, which can be computationally expensive. To address this, we aim to integrate finger structural information, such as joint angles, into the contact points, thus eliminating the need for IK computation.

We represent the workspace of the fingertips using point clouds, where each point corresponds to the position of a fingertip contact point in the palm frame \(frame_{palm}\) during a grasp. By leveraging the flexible structure of point clouds, we extend the fields to include additional information beyond position and normal with corresponding joint angles and contact surface properties. This allows for the direct retrieval of joint angles when a point is selected. The AutoWS generation process of workspace clouds is illustrated in Fig.\ref{fig:autows}.

\subsection{Fingertip Contact Face Approximation}

From the robot model, we obtain the mesh of the fingertip links. PDS \cite{poisson} is applied to generate the corresponding clouds, capturing both the surface point positions and normal directions.

Due to the non-uniform orientation of fingertip meshes, two directional vectors are required as input: a forward vector \(v_{f}\), pointing toward the pulp side, and an upward vector \(v_{u}\), aligned with the finger in the tip frame \(frame_{tip}\). To avoid generating grasp poses using the fingernail side, we establish a viewpoint at a distance along the fingertip up-forward direction, relative to the mesh centroid \(cent\). The viewpoint coordinate is calculated as \(co_{viewpoint} = cent + 100 \times d \times v_{f} + 50 \times d \times v_{u}\), where \(d\) represents the diagonal length of the mesh bounding box. Following the method from \cite{hidden}, we remove points that are hidden from the viewpoint perspective. The remaining points in cloud, denoted as \(cloud_{f}\) in Fig.\ref{fig:autows}(b), are considered candidates of contact face center.

For a demonstrated grasp pose, as shown in Fig.\ref{fig:autows}(c), the centroid of all actual contact points on each fingertip is used as the initial query point \(q_{init}\). Using a kd-tree, we find the nearest point in \(cloud_{f}\), which serves as the center point \(c\) of the disired contact face. In the absence of demonstrations, the centroid of \(cloud_{f}\) serves as \(q_{init}\) for nearest point searching.

Denote \(pos_{c}\), \(nor_{c}\), \(pos_{p}\) and \(nor_{p}\) as the position and normal vectors of the center point \(c\) and other points \(p\) in \(cloud_{f}\). Points that are close to \(c\) along \(nor_{c}\) and have similar normal directions, satisfying the conditions in \eqref{eq:proj}, are projected onto a plane perpendicular to \(nor_{c}\), with thresholds \(thre_{pos}\) and \(thre_{nor}\). The projected points form a convex hull, and the largest circle or axis-aligned rectangle within the hull is designated as the corresponding contact face, as shown in Fig.\ref{fig:autows}(f).

\begin{equation}
    \begin{aligned}
     \left\| \left< pos_{p} - pos_{c},  nor_{c} \right>  \right\| > thre_{pos} \\
     \left< nor_{p} , nor_{c}\right> > thre_{nor}
    \label{eq:proj}
    \end{aligned}
\end{equation}

To cover more of the fingertip surface and retain more flexibility in contact point selection, we select three additional points located on the upper, left and right sides of the initially formed contact face, as shown in Fig.\ref{fig:autows}(g) for the upper side condition. These points are utilized as centers for reconstructing new contact faces. If the area of these contact faces is less than \(1/4\) of the first contact face, the corresponding contact face is discarded.

\subsection{Workspace Cloud Generation and Preprocessing} \label{sec:ws_prepro}

We unify the hand orientation by aligning the palm with the \(X\text{-}O\text{-}Z\) plane, as shown in Fig.\ref{fig:autows}(a), and generate fingertip workspace clouds in \(frame_{palm}\).

The contact point candidate position and normal are computed by sampling within the joint angle space. When human demonstration data is available, the joint angles from the demonstrated grasp gesture serve as the centers of the target angle ranges for the corresponding joints. The joint angles are incrementally adjusted within their respective physical ranges, using configurable step sizes and counts, to generate a series of angle combinations. In this work, the step size is set to five degrees, with two steps for both increasing and decreasing. When no human demonstration is available, the target joint angle ranges and step sizes have to be specified manually. Using the position and normal of the center point of each approximated contact face in \(frame_{tip}\), forward kinematics is applied to compute the actual position and normal direction of these center points for each angle combination in \(frame_{palm}\). This process forms the initial fingertip workspace clouds.

For each point, we store the position, normal direction, joint angles required for the fingertip to reach that point, and properties of the contact face shape. For rectangular shapes, we store vectors along two perpendicular edges, with their norms indicating half the edge lengths. For circular shapes, we store the radius.

To accelerate subsequent contact point sampling, we filter out invalid points from the original clouds in Fig.\ref{fig:autows}(h). For the workspace cloud of each finger, the average normal vector direction \(nor_{ave}\) is computed. If the y-component of \(nor_{ave}\) is positive, indicating the fingertip is primarily facing the palm, all points in the cloud with a normal y-component less than -0.001 are removed. Conversely, if the y-component of \(nor_{ave}\) is non-positive, all points with a normal y-component greater than 0.001 are removed. Fig.\ref{fig:autows}(i) shows the fingertip workspace clouds after filtering.

For all valid points, additional point clouds, \(cloud_{link}\), are generated for the skeletons of links other than the fingertips. These clouds, as shown in Fig.\ref{fig:autows}(j), are subsequently used to detect potential penetrations between the links and the object being grasped.

The generated fingertip workspace clouds are manually ranked in order of priority, based on the importance of each finger in the grasping task. For a dexterous hand, the thumb is considered the most important, while the little finger is ranked the least important.
}
{
}

\begin{figure}
  \centering
  \includegraphics[width=\columnwidth]{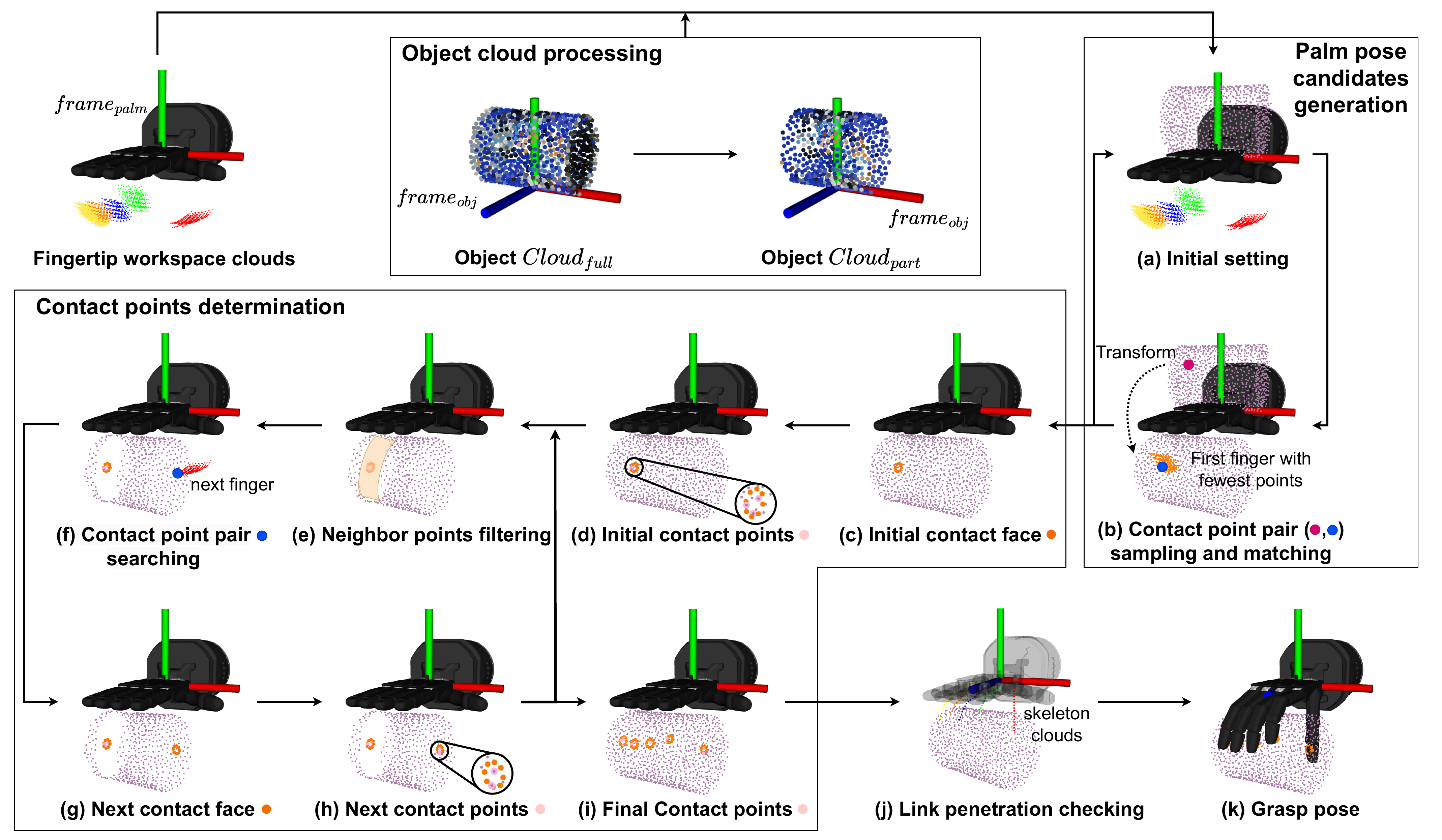}
  \caption{Flowchart of FSG. Fingertip workspace clouds and object clouds are inputs to FSG. (a) Temporarily align \(frame_{obj}\) with \(frame_{palm}\). (b) Generate palm pose candidates by random sampling and matching contact point pairs between \(cloud_{part}\) and the workspace cloud with least points. (c) Reconstruct fingertip contact face. (d) Find contact points from \(cloud_{full}\) located within fingertip contact face. (e) Filter out neighbor points from \(cloud_{part}\). (f-h) Iteratively search for contact faces and points of other fingers. (i) Final contact points from \(cloud_{full}\) located within all fingertip contact faces. (j) Get hand skeleton clouds based on final contact points. (k) Generated grasp pose for further quality evaluation.
  }
  \label{fig:ranspreg}
\end{figure}

\section{Fingertip-Contact-Aware Sampling-based Grasp Generator}

\switchlanguage%
{

To obtain the three key components of a precision grasp pose, it is required to first determine the relative pose between the palm and the object, and then identify suitable contact points in the workspace clouds and their corresponding joint angles. In this section, we present a fingertip-contact-aware sampling-based approach that efficiently determines the relative palm pose and contact point information. The overall workflow is shown in the Fig.\ref{fig:ranspreg}.

\subsection{Object Cloud Processing} \label{obj_proc}
We represent the object shape using a point cloud and preprocess it to remove regions unsuitable for grasping. For the complete object cloud, \(cloud_{full}\), we annotate and extract the target graspable regions based on affordance information. If affordance annotations are unavailable, we estimate the curvature at each point based on \cite{curv} and exclude points with relatively high curvature, using the remaining regions as the target graspable part, denoted as \(cloud_{part}\).

To determine the relative pose between the object and the palm, we first establish the object frame, \(frame_{obj}\). When generating grasp data for floating objects, \(frame_{obj}\) can be arbitrarily assigned. However, for objects resting on a table, we align the \(X\text{-}O\text{-}Z\) plane of \(frame_{obj}\) with the table surface. This alignment simplifies subsequent collision checking between the hand and the table.

\subsection{Grasp Synthesis}

\subsubsection{Palm Pose Candidates Generation} \label{palm_cand}

Each point in the workspace cloud generated in Sec.\ref{sec:ws_gen} represents a potential contact point from the corresponding fingertip. By randomly sampling a point from the object cloud and another from the workspace cloud, these two points form a contact point pair. Relative poses between the object and the palm, i.e., the palm pose candidates relative to \(frame_{obj}\), are identified by matching the point pair in a dual-loop strategy.

In the outer loop, we sample one point from \(cloud_{part}\) and another point from the workspace cloud with fewest points. \(cloud_{part}\) is then transformed so that the sampled point pair coincides in both position and normal direction, as shown in Fig.\ref{fig:ranspreg}(b). When sampling a contact point from a workspace cloud, we reconstruct the contact face and uniformly sample a point from it as the final contact point for that finger. The normal direction at the center point is used to approximate the normal directions of all sampled points from that contact face.

Since the position and normal direction of a single point pair are insufficient to determine the relative pose, we introduce an inner loop where \(cloud_{part}\) is rotated around the normal of the aligned contact point pair. This rotation process not only determines the relative pose but also generates additional palm pose candidates.

The same transformation is applied to \(cloud_{full}\). For precision grasp, we filter out cases where the transformed \(cloud_{full}\) either penetrates the palm or collides with it. This is achieved by counting the number of object points within the palm bounding box. For scenarios involving palm-table collision detection, we take advantage of the fact that the \(X\text{-}O\text{-}Z\) plane of \(frame_{obj}\) represents the table surface. By computing the height of the palm bounding box vertices relative to the table, we filter out cases where the palm height is lower than a predefined threshold.

\subsubsection{Contact Points Determination} \label{sec:contact}

After obtaining the palm pose candidates, we proceed to determine the contact points for each fingertip. The first contact point is derived from the contact point pair sampling process described in Sec.\ref{palm_cand}. For each of the remaining fingers, we construct a k-d tree for the corresponding workspace cloud and search for a pair of points \((p_{obj}, p_{fin})\) that are sufficiently close in both position and normal direction between the transformed \(cloud_{part}\) and the fingertip workspace cloud.

For each identified contact point pair, we filter out points from the transformed \(cloud_{part}\), whose position \(x_{point}\) along the \(x\)-axis of \(frame_{palm}\) and normal \(nor_{point}\) satisfy the conditions in \eqref{eq:filt}, before proceeding to search for the next pair, as shown in Fig.\ref{fig:ranspreg}(e).

\begin{equation}
    \begin{aligned}
     x_{point} \in \left [ x_{con}-w_{tip} \times \left\| nor_{x}\right\|/\sqrt{nor_{x}^{2} + nor_{z}^{2}}, x_{con}+w_{tip} \times \left\| nor_{x}\right\|/\sqrt{nor_{x}^{2} + nor_{z}^{2}} \right ] \\
     \left< nor_{point} , nor_{con}\right> > 0
    \label{eq:filt}
    \end{aligned}
\end{equation}

where \(x_{con}\) is the position of \(p_{obj}\) along \(x\)-axis, \(nor_{x}\) and \(nor_{z}\) are the \(x-\) and \(z\)-axis components of the normal vector at \(p_{obj}\), respectively, and \(w_{tip}\) is the width of the fingertip. This filtering ensures that subsequent contact points for other fingers are not selected in nearby regions, thus preventing finger collisions.

Workspace clouds are divided into two groups based on the opposition of their own average normals. Within each group, the workspace clouds are sorted by their average position along the \(y\)-axis of \(frame_{palm}\). If the order of the identified contact points along the \(y\)-axis does not match the order of the corresponding workspace clouds, those contact points are discarded to prevent finger crossing.

Using the contact face information stored in each finger contact point \(p_{fin}\), we reconstruct the corresponding contact face, as shown in Fig.\ref{fig:ranspreg}(c) and Fig.\ref{fig:ranspreg}(g). For finger-table collision detection, we calculate the distance from each vertex of the reconstructed contact face to the table surface (the \(X\text{-}O\text{-}Z\) plane of \(frame_{obj}\)) and compare it to a threshold to identify collisions. These reconstructed contact faces are then used to find points from \(cloud_{full}\) that lie within the contact face and are close to the face along its normal direction, as shown in Fig.\ref{fig:ranspreg}(d) and Fig.\ref{fig:ranspreg}(h). These points are selected as the final contact points.

\subsubsection{Grasp Quality Evaluation} \label{sec:quality}

We use the quality measure \(Q_{1}\), also known as grasp wrench space (GWS) epsilon, as proposed in \cite{gws}, to evaluate the quality of grasp pose candidates by treating all fingertips as rigid bodies.

At each contact point, we assume a unit force is applied in the direction of the contact normal. This force, combined with friction, forms a friction cone at the contact point. For computational simplicity, the friction cone is approximated by an octahedron, and the wrench applied at the contact point is represented as a linear combination of the maximal wrenches along the edges of the octahedron. These wrenches are then transformed from the contact point frame to the object centroid frame. The total wrench space, represented using the \(L_{1}\) metric, corresponds to the convex hull formed by all individual wrenches, as shown in Eq.\eqref{eq:gws}.

\begin{equation}
    W_{L_{1}} = ConvexHull( \bigcup_{i=1}^{n} \{w_{i,1}, ..., w_{i,m}\})
    \label{eq:gws}
\end{equation}

where n is the number of contact points and m is the number of edges in the friction cone approximation. If the origin lies inside the convex hull, the grasp pose is considered to be in force closure. The distance from the origin to the nearest facet of the convex hull is denoted as \(Q_{1}\). A larger positive value of \(Q_{1}\) indicates a grasp pose with better resistance to external force disturbances, while a negative \(Q_{1}\) indicates that force closure is not satisfied.

For face contact, the vertices of the contact face are typically used as the contact points. However, since both the fingertips and the object are represented using point clouds, precisely determining the actual contact face is challenging. Instead, we directly use all contact points found on the object surface, as described in Sec.\ref{sec:contact}, for computation. The computed \(Q_{1}\) is then used to filter and rank the grasp pose candidates.

\subsubsection{Finger Iteration} \label{sec:finIter}

Each finger operates within its own independent workspace cloud, allowing for control over the number of fingers or specific fingers used in grasping by activating the corresponding clouds. Initially, We assume that all fingers are required to grasp the object. However, if the number of valid gestures is insufficient, we relax this condition by reducing the number of fingers required for the grasp, thus enabling the identification of additional grasp poses. Fingers are disabled starting with the least important, following the priority sequence outlined in Sec.\ref{sec:ws_prepro}. For the unused fingers, their joint angles are set to 0 in the grasp pose.

\subsubsection{Link Penetration Checking}

In the grasp pose generation process, we focused solely on the relationship between the palm, fingertips, and the object, while neglecting the other links of the fingers. To prevent penetration between the object and the other finger links, we identify points in \(cloud_{link}\) that represents the skeletons of the other finger links, based on the joint angles corresponding to each contact point, as shown in Fig.\ref{fig:ranspreg}(j). Starting from one end of each finger link skeleton, we sequentially search for the nearest points in the transformed \(cloud_{obj}\). For adjacent pairs of 3D positions along the skeleton, \(p_{n}\) and \(p_{n+1}\), and their nearest points on the object, \(q_{n}\) and \(q_{n+1}\), with the object facial normal stored with \(q_{n}\) being \(nor_{n}\), if they satisfy the condition in \eqref{eq:link}, we consider penetration to have occurred in that section and discard the corresponding grasp pose.

\begin{equation}
    [(p_{n} - q_{n}) \cdot (nor_{n})] \times [(p_{n+1} - q_{n+1}) \cdot (nor_{n})] < 0
    \label{eq:link}
\end{equation}

}
{
}

\section{Experiments}

\switchlanguage%
{

In this section, we use objects from the YCB dataset \cite{ycb} as grasping targets to evaluate the performance of our method in data augmentation with the Shadow Hand \cite{shadow}. Additionally, we test the capability of our method to generate grasp pose directly, without relying on human demonstration data. We also validate its generation capability on robotic hands with varying numbers of fingers and different mechanical structures.

To acquire human demonstration data, we use the Meta Quest to track hand postures and teleoperate the Shadow Hand in the MuJoCo simulator \cite{mujoco} to perform grasps. The palm of the Shadow Hand follows the movement of the demonstrator's palm, while the joint angles are derived from the hand posture detected by the Meta Quest and mapped onto the Shadow Hand to control finger flexion. In the simulator, we add a camera following the movements of the headset, rendering real-time images to the headset to provide visual feedback. To compensate for the lack of tactile feedback, we display contact points directly in the rendered images. It takes us around three hours to collect demonstration data for 74 YCB objects.

For the target objects, we position them above a table, allow them to fall and stabilize, and then fix their poses. During the grasp pose demonstration, we record the joint angles of the fingers and the relative pose of the palm with respect to the object. These data, combined with the finger contact points, are used to generate the corresponding workspace clouds and link skeleton clouds. The object cloud \(cloud_{obj}\) is obtained by sampling 1,024 points from the object mesh surface using PDS, and the target grasp part of each object, \(cloud_{part}\), is manually annotated.

We compare our method against baselines including \(EigenGrasp\) \cite{eigen}, \(DexGraspNet\) \cite{dex}, and \(QD\text{-}Grasp\text{-}6DoF\) \cite{qd}, evaluating grasping performance on the YCB objects with the Shadow Hand. \(DexGraspNet\) is tested on an NVIDIA RTX 4090 GPU, while our method, \(EigenGrasp\), and \(QD\text{-}Grasp\text{-}6DoF\) are tested on Intel(R) Xeon(R) Gold 6336Y CPU @ 2.40GHz. Both our method and \(QD\text{-}Grasp\text{-}6DoF\) support multi-threaded parallel processing, and we utilize 12 threads for execution. In our method, the outer loop is set to 200 iterations, and the inner loop is set to 15 iterations for generating the palm pose candidates.

\subsection{Hand-Only Grasp Data Generation} \label{sec:gdg}

In this experiment, we focus solely on the interaction between the robotic hand and the object. To simplify the setup, the objects are suspended in the air, eliminating the need to consider collisions between the robotic hand and the surrounding environment. Grasp poses are generated directly under these conditions.

We first evaluate the performance of our proposed method for generating grasp data without relying on human demonstration. In the absence of demonstration , we generate the fingertip workspace clouds using AutoWS, based on the manually defined joint range and step size for each finger joint. The angle ranges for the finger joints are listed in Table\ref{table:range}, where FJ1 represents the distal joint. Since the four fingers, excluding the thumb, are underactuated in the Shadow Hand, we set the joint angle of FJ1 to match that of FJ2 during workspace generation. All angle step sizes are set to 10 degrees.

\begin{table}[h]
  \centering
  \caption{Angle ranges of finger joints for Shadow Hand workspace cloud generation (degree)}
  \scalebox{0.7}{
  \begin{tabular}{|c|c|c|c|c|c|c|c|c|c|}
    \hline
    Forefinger & Range & Middle finger & Range & Ring finger & Range & Little finger & Range & Thumb & Range \\
    \hline
    FFJ4 & [-10, 10] & MFJ4 & [-10, 10] & RFJ4 & [-10, 10] & LFJ5 & [  0,  0] & THJ5 & [10,30] \\
    FFJ3 & [ 20, 50] & MFJ3 & [ 20, 50] & RFJ3 & [ 20, 50] & LFJ4 & [-10, 10] & THJ4 & [50,70] \\
    FFJ2 & [ 10, 40] & MFJ2 & [ 10, 40] & RFJ2 & [ 10, 40] & LFJ3 & [ 20, 50] & THJ3 & [-10,10] \\
    FFJ1 & [ 10, 40] & MFJ1 & [ 10, 40] & RFJ1 & [ 10, 40] & LFJ2 & [ 10, 40] & THJ2 & [-20,20] \\
         &           &      &           &      &           & LFJ1 & [ 10, 40] & THJ1 & [-10,30] \\
    \hline
  \end{tabular}}
  \label{table:range}
\end{table}

For comparison, we test all baselines with their default settings. Specifically, for \(QD\text{-}Grasp\text{-}6DoF\), we use the \(contact\_me\_scs\) algorithm to generate grasp poses. In our method, we set a threshold of 10 generated poses per execution. If the number of poses is insufficient and finger iteration is triggered, we reduce the number of fingers and continue generating poses within the same execution until the required number is reached. The comparative experiments are conducted on five YCB objects without scaling. To standardize the evaluation of grasp quality across different methods, we import the generated grasp poses into GraspIt!\cite{graspit} and calculate the grasp quality using the GWS epsilon metric and GWS volume metric. The results are summarized in Table\ref{table:gdg}.

In Table\ref{table:gdg}, we use \(FSG\) to represent our approach, where a circular contact surface is used for generating workspace clouds, and the finger iteration is applied by default. \(FSG\text{+}Aff\) builds upon FSG by utilizing the point cloud of the target grasp part based on affordance, rather than the entire object cloud. \(FSG\text{+}Aff\text{-}FinIter\) remove the finger iteration module of \(FSG\text{+}Aff\). \(FSG\text{+}Aff\text{+}LinkPen\) incorporates link penetration checking into \(FSG\text{+}Aff\).

For grasping tasks involving relatively simple shapes such as the 004\text{-}sugar\text{-}box, 005\text{-}tomato\text{-}soup\text{-}can, and 043\text{-}phillips\text{-}screwdriver, all methods can generate corresponding grasp poses. Among them, our approach has an advantage in terms of generation speed. Comparing \(FSG\text{+}Aff\) with \(FSG\), using the target grasp part cloud reduces the number of points, making it easier to sample suitable contact positions during contact point pair sampling. This increases the probability of generating a valid grasp pose in each iteration, thus speeding up pose generation. When comparing \(FSG\text{+}Aff\text{-}FinIter\) with \(FSG\text{+}Aff\), removing finger iteration requires all fingers to make contact with the object to complete the grasp, which is more demanding and slows down the generation speed. Comparing \(FSG\text{+}Aff\text{+}LinkPen\) with \(FSG\text{+}Aff\), adding link penetration detection after grasp pose generation increases computational cost. However, since link penetration checking is only performed on grasp poses that meet the contact points and \(Q_{1}\) conditions, whose number is relatively small, the additional time cost is minimal.

In GraspIt, when calculating the GWS epsilon, it is necessary to first determine the contact points. Meanwhile, even slight penetration between the fingers and the object can cause the calculation to fail. Therefore, for all input grasp poses, we first slightly open the fingers based on the grasp pose, then sequentially close each finger to find the corresponding contact points. During the finger closing process, if there is a collision between fingers, it becomes difficult to obtain the corresponding contact points, which can also lead to a failure in the epsilon calculation. The GWS epsilon and GWS volume metrics in Table\ref{table:gdg} show the average values of epsilon and volume for all grasp poses determined by GraspIt to be force-closure. There is no significant difference on two metrics between all methods. However, our approach performs relatively better in terms of the valid epsilon percentage.

For the 024\text{-}bowl, which requires grasping the edge, and the 062\text{-}dice, which is very small, although all three baselines can generate grasp poses, the valid epsilon percentage is nearly zero. Due to the thin and small nature of the grasping part of these two objects, the valid fingertip contact point combinations exceed the workspace clouds generated by the specified joint ranges, which makes it impossible for our approach to generate valid grasp poses.

\begin{table}[h!]
\centering
\caption{Performance on hand-only grasp data generation}
\label{table:gdg}
\resizebox{0.8\textwidth}{!}{%
\renewcommand{\arraystretch}{1.3}
\begin{tabular}{|c|c|c|c|c|c|}
\hline
Object                   & Method                  & Time Cost / Result (s) $\downarrow$ & GWS Epsilon $\uparrow$  & GWS Volume $\uparrow$    & Valid Epsilon Percentage $\uparrow$ \\ \hline
\multirow{8}{*}{004-sugar-box} 
& EigenGrasp              & 1.622                       & $9.16 \text{e-}03$                       & $8.96 \text{e-}04$                       & 0.3\%                        \\
& DexGraspNet             & 0.525                       & $1.62 \text{e-}02$                       & $2.40 \text{e-}03$                       & 1.5\%                        \\
& QD-Grasp-6DoF           & 4.225                       & $9.58 \text{e-}03$                       & $1.74 \text{e-}03$                       & 6.4\%                        \\ \cline{2-6} 
& FSG                     & 0.147                       & $1.80 \text{e-}02$                       & $2.13 \text{e-}03$                       & 18.6\%                       \\
& FSG+Aff                 & \underline{\textbf{0.077}}  & $2.10 \text{e-}02$                       & $2.37 \text{e-}03$                       & \underline{\textbf{19.1\%}}  \\
& FSG+Aff-FinIter         & 0.187                       & \underline{$\mathbf{2.18 \text{e-}02}$}  & \underline{$\mathbf{3.82 \text{e-}03}$}  & 16.4\%                       \\
& FSG+Aff+LinkPen         & 0.087                       & $1.89 \text{e-}02$                       & $2.18 \text{e-}03$                       & 17.7\%                       \\ \hline \hline

\multirow{8}{*}{005-tomato-soup-can} 
& EigenGrasp              & 1.619                       & $2.32 \text{e-}02$                       & $8.52 \text{e-}03$                       & 0.1\%                        \\
& DexGraspNet             & 0.525                       & $1.94 \text{e-}02$                       & \underline{$\mathbf{1.09 \text{e-}02}$}  & 5.9\%                        \\
& QD-Grasp-6DoF           & 13.847                      & $2.07 \text{e-}02$                       & $8.48 \text{e-}03$                       & 36.5\%                       \\ \cline{2-6} 
& FSG                     & 0.207                       & $2.90 \text{e-}02$                       & $1.05 \text{e-}02$                       & 34.9\%                       \\
& FSG+Aff                 & \underline{\textbf{0.056}}  & $2.64 \text{e-}02$                       & $5.37 \text{e-}03$                       & 60.8\%                       \\
& FSG+Aff-FinIter         & 0.541                       & \underline{$\mathbf{3.29 \text{e-}02}$}  & $1.01 \text{e-}02$                       & \underline{\textbf{86.8\%}}  \\
& FSG+Aff+LinkPen         & 0.066                       & $2.57 \text{e-}02$                       & $5.32 \text{e-}03$                       & 71.9\%                       \\ \hline \hline

\multirow{8}{*}{024-bowl} 
& EigenGrasp              & 1.621                       & NaN                                      & NaN                                      & 0.0\%                       \\
& DexGraspNet             & 0.525                       & \underline{$\mathbf{1.91 \text{e-}02}$}  & $2.91 \text{e-}03$                       & 0.1\%                       \\
& QD-Grasp-6DoF           & \underline{\textbf{0.297}}  & $6.59 \text{e-}03$                       & \underline{$\mathbf{3.11 \text{e-}03}$}  & \underline{\textbf{0.3\%}}  \\ \cline{2-6} 
& FSG                     & NaN                         & NaN                                      & NaN                                      & NaN                         \\
& FSG+Aff                 & NaN                         & NaN                                      & NaN                                      & NaN                         \\
& FSG+Aff-FinIter         & NaN                         & NaN                                      & NaN                                      & NaN                         \\
& FSG+Aff+LinkPen         & NaN                         & NaN                                      & NaN                                      & NaN                         \\ \hline \hline

\multirow{8}{*}{043-phillips-screwdriver} 
& EigenGrasp              & 1.614                       & \underline{$\mathbf{1.78 \text{e-}02}$}  & $5.39 \text{e-}04$                       & 0.3\%                        \\
& DexGraspNet             & 0.525                       & $2.65 \text{e-}03$                       & $1.79 \text{e-}04$                       & 0.2\%                        \\
& QD-Grasp-6DoF           & 0.579                       & $5.40 \text{e-}03$                       & $5.25 \text{e-}04$                       & 2.2\%                        \\ \cline{2-6} 
& FSG                     & 0.181                       & $8.41 \text{e-}03$                       & $4.38 \text{e-}04$                       & 18.0\%                       \\
& FSG+Aff                 & 0.128                       & $6.38 \text{e-}03$                       & $1.63 \text{e-}04$                       & 36.2\%                       \\
& FSG+Aff-FinIter         & 2.536                       & $1.22 \text{e-}02$                       & \underline{$\mathbf{7.58 \text{e-}04}$}  & \underline{\textbf{45.0}}\%  \\
& FSG+Aff+LinkPen         & \underline{\textbf{0.120}}  & $7.44 \text{e-}03$                       & $2.19 \text{e-}04$                       & 39.3\%                       \\ \hline \hline

\multirow{8}{*}{062-dice} 
& EigenGrasp              & 1.624                       & NaN                    & NaN                & 0.0\%               \\
& DexGraspNet             & \underline{\textbf{0.525}}  & NaN                    & NaN                & 0.0\%               \\
& QD-Grasp-6DoF           & 2.719                       & NaN                    & NaN                & 0.0\%               \\ \cline{2-6} 
& FSG                     & NaN                         & NaN                    & NaN                & NaN                 \\
& FSG+Aff                 & NaN                         & NaN                    & NaN                & NaN                 \\
& FSG+Aff-FinIter         & NaN                         & NaN                    & NaN                & NaN                 \\
& FSG+Aff+LinkPen         & NaN                         & NaN                    & NaN                & NaN                 \\ \hline

\end{tabular}
}
\end{table}

Apart from experimental comparisons, our approach enables direct specification of the number of fingers used for grasping through the finger iteration functionality. Grasp poses generated with different numbers of fingers for grasping the YCB 021\text{-}bleach\text{-}cleanser are shown in Fig.\ref{fig:finIter_example}. Additionally, since our method implicitly incorporates the structural information of the robotic hand into workspace clouds, the grasp generation process does not depend on the specific structure of the robotic hand. As long as the fingertip workspace clouds are provided, FSG can be applied to robotic hands of any structure. Generated grasps for the Robotiq 2F-85 gripper \cite{robotiq}, Barrett Hand \cite{barrett}, and Allegro Hand \cite{allegro} are shown in Fig.\ref{fig:whole}.

\begin{figure}
  \centering
  \includegraphics[width=0.9\columnwidth]{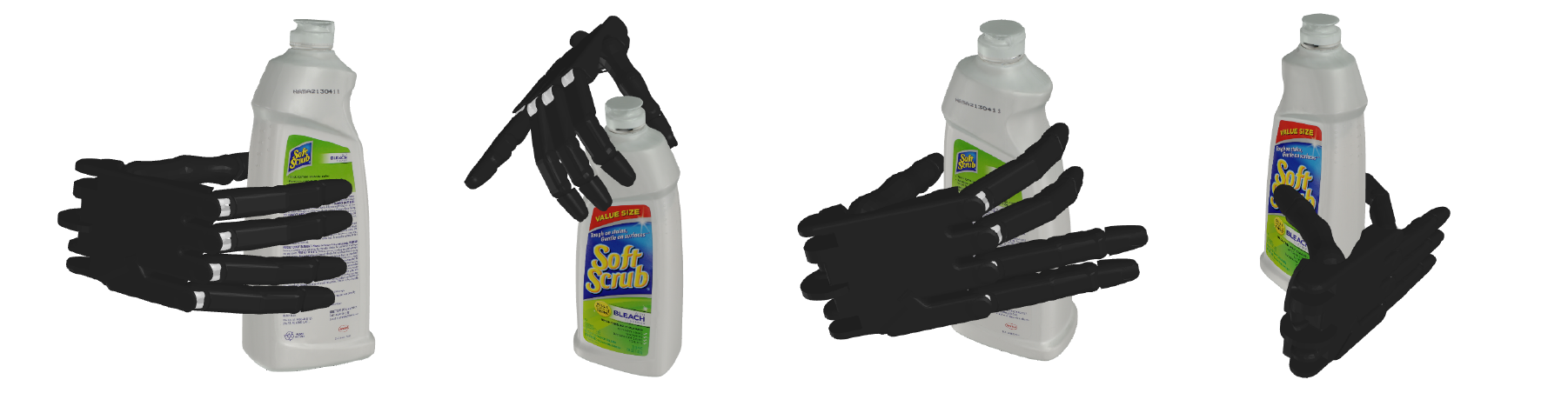}
  \caption{Poses generated for the Shadow Hand to grasp the ycb 021\text{-}bleach\text{-}cleanser with different number of fingers.}
  \label{fig:finIter_example}
\end{figure}

\subsection{Hand-Only Grasp Data Augmentation} \label{sec:gda}
In this section, we evaluate the performance of our proposed method for data augmentation on all YCB objects with complete mesh models. For each object, we collect one grasp demonstration via teleoperation. Use the joint angles and contact points from the demonstration, we generate the corresponding fingertip workspace clouds.

The performance of our method, after incorporating the grasp pose demonstration data, is summarized in Table \ref{table:gda} for the five YCB objects tested in Sec.\ref{sec:gdg}.

When generating the workspace clouds based on a grasp pose demonstration, the joint angle ranges are set covering the joint angles of the demonstration. This ensures that valid fingertip contact point combinations lie within the generated workspace cloud. Therefore, for the 024-bowl and 062-dice, which were challenging to grasp in Sec.\ref{sec:gdg}, invoking FSG with the updated workspace clouds enables rapid grasp generation.

In terms of generation speed, using the object target grasp part cloud continues to accelerate the generation process. Additionally, as shown in Table\ref{table:gda}, the speed of \(FSG\text{+}Aff\text{-}FinIter\) is very close to, or even faster than, \(FSG\text{+}Aff\). This is primarily because, with the newly generated workspace clouds, \(FSG\text{+}Aff\) can, in most cases, generate sufficient grasp poses in a single execution without triggering finger iteration. This minimizes the difference between \(FSG\text{+}Aff\text{-}FinIter\) and \(FSG\text{+}Aff\). When link penetration checking is included, a significant decrease in speed is observed for the 024-bowl. This slowdown is due to the fact that, without link penetration checking, some invalid grasp poses are mistakenly considered valid, leading to an overestimation of the number of valid grasps. In contrast, for other objects, the inclusion of link penetration checking does not cause a similar issue. This is because these objects do not feature large concave surfaces, while concave surfaces could lead to grasp poses with perfect contact points but finger penetration, as shown in Fig.\ref{fig:bowl_pen}.

In the case of 062-dice, the evaluated valid epsilon percentage in GraspIt remains very low. Since only two fingers are used in the demonstration, the generated grasp poses involve only those two fingers. As a result, when determining the collision points between the object and each fingertip, typically only one contact point is found per finger, resulting in a total of just two contact points. This limitation directly causes the failure in calculating the GWS epsilon, with the requirement of at least three contact points.

In the case of 024-bowl, almost no valid epsilon values are found during evaluation in GraspIt. Two common failure scenarios are illustrated in Fig.\ref{fig:bowl_fail}. The first failure (on the left) occurs due to slight penetration between the bowl edge and the middle finger, which directly hinders proper finger closure and prevents GWS epsilon calculation. Although our method includes penetration checking to avoid fingers passing through the object surface, minor penetrations remain challenging to handle effectively. The second failure (on the right) arises from issues with contact point determination. Since fingertips are treated as solid, the estimated contact points may exhibit unexpected orientations, even in an antipodal configuration, leading to an invalid GWS epsilon.

To improve the valid epsilon percentage, potential solutions include implementing more dedicated penetration checking and using soft fingertip models for contact point determination. These enhancements are left for future work.

\begin{figure}
    \centering
    \begin{minipage}{0.45\textwidth}  
        \centering
        \includegraphics[height=4cm]{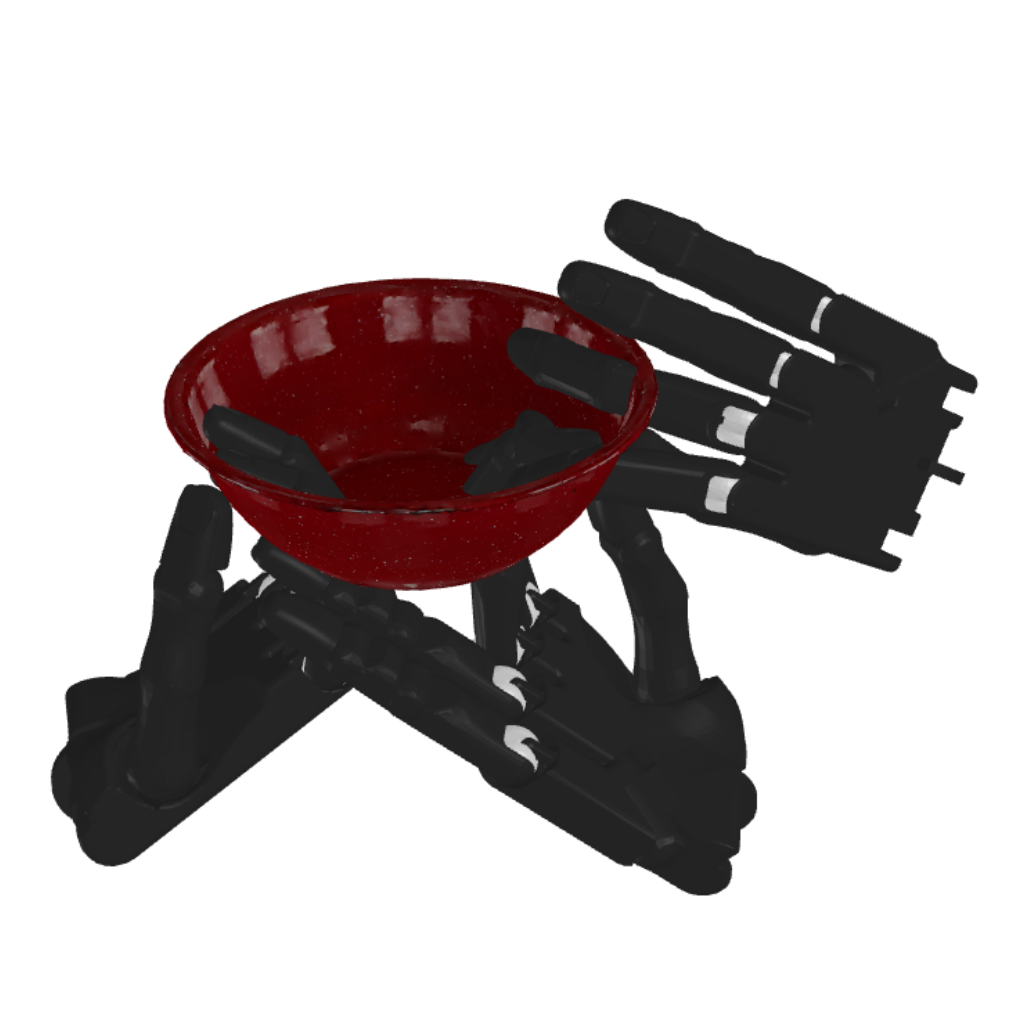} 
        \caption{Grasp poses on concave surface without link penetration checking.}
        \label{fig:bowl_pen}
    \end{minipage}%
    \hfill 
    \begin{minipage}{0.45\textwidth}  
        \centering
        \includegraphics[height=4cm]{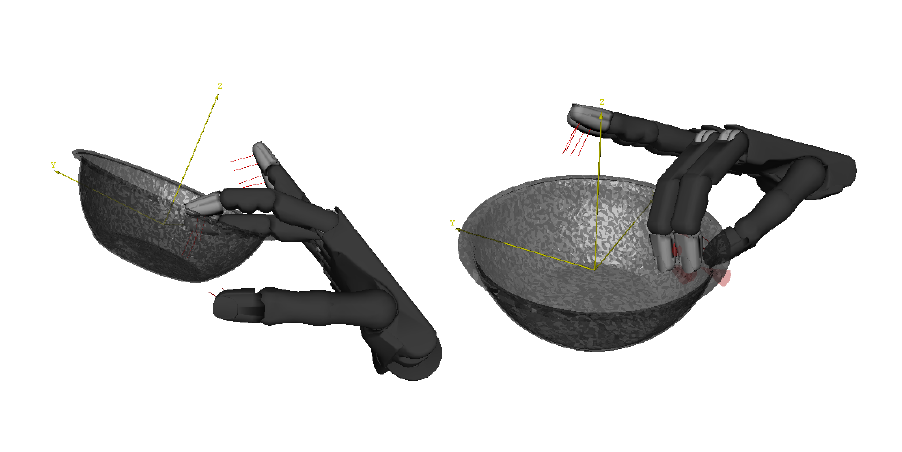} 
        \caption{Grasp poses with invalid GWS epsilons evaluated in graspit.}
        \label{fig:bowl_fail}
    \end{minipage}
\end{figure}

\begin{table}[h!]
\centering
\caption{Performance on hand-only grasp data augmentation}
\label{table:gda}
\resizebox{0.8\textwidth}{!}{%
\renewcommand{\arraystretch}{1.3}
\begin{tabular}{|c|c|c|c|c|c|}
\hline
Object                   & Method                  & Time Cost / Result (s) $\downarrow$ & GWS Epsilon $\uparrow$  & GWS Volume $\uparrow$    & Valid Epsilon Percentage $\uparrow$ \\ \hline

\multirow{4}{*}{004-sugar-box} 
& FSG               & 0.138                      & 1.54$\text{e-02}$                      & 1.04$\text{e-03}$                      & 41.8\%                      \\
& FSG+Aff           & 0.088                      & 1.60$\text{e-02}$                      & 1.28$\text{e-03}$                      & \textbf{\underline{47.9\%}} \\
& FSG+Aff-FinIter   & \underline{\textbf{0.086}} & \textbf{\underline{1.65$\text{e-02}$}} & \textbf{\underline{1.36$\text{e-03}$}} & 42.7\%                      \\
& FSG+Aff+LinkPen   & 0.087                      & 1.59$\text{e-02}$                      & 1.24$\text{e-03}$                      & 46.3\%                      \\ \hline \hline

\multirow{4}{*}{005-tomato-soup-can}
& FSG               & 0.053                      & 2.98$\text{e-02}$                      & \textbf{\underline{6.20$\text{e-03}$}} & 55.9\%                      \\
& FSG+Aff           & 0.042                      & 2.88$\text{e-02}$                      & 6.00$\text{e-03}$                      & 60.5\%                      \\
& FSG+Aff-FinIter   & \underline{\textbf{0.041}} & 2.98$\text{e-02}$                      & 5.72$\text{e-03}$                      & 62.6\%                      \\
& FSG+Aff+LinkPen   & 0.043                      & \textbf{\underline{3.28$\text{e-02}$}} & 5.95$\text{e-03}$                      & \textbf{\underline{64.9\%}} \\ \hline \hline

\multirow{4}{*}{024-bowl}
& FSG               & 0.100                      & 5.43$\text{e-03}$                      & 5.04$\text{e-04}$                      & 0.2\%                      \\
& FSG+Aff           & 0.039                      & 7.42$\text{e-03}$                      & 2.36$\text{e-04}$                      & 0.4\%                      \\
& FSG+Aff-FinIter   & \textbf{\underline{0.036}} & \textbf{\underline{2.00$\text{e-02}$}} & \textbf{\underline{1.27$\text{e-03}$}} & 0.4\%                      \\
& FSG+Aff+LinkPen   & 0.087                      & 7.28$\text{e-03}$                      & 5.69$\text{e-04}$                      & \textbf{\underline{0.8\%}} \\ \hline \hline

\multirow{4}{*}{043-phillips-screwdriver}
& FSG               & 0.161                      & \underline{\textbf{1.51$\text{e-02}$}} & 7.61$\text{e-04}$                      & 52.2\% \\
& FSG+Aff           & \underline{\textbf{0.143}} & 1.37$\text{e-02}$                      & 6.00$\text{e-04}$                      & 51.8\% \\
& FSG+Aff-FinIter   & 0.189                      & 1.47$\text{e-02}$                      & \underline{\textbf{7.85$\text{e-04}$}} & \underline{\textbf{52.9\%}} \\
& FSG+Aff+LinkPen   & 0.168                      & 1.29$\text{e-02}$                      & 5.35$\text{e-04}$                      & 52.4\% \\ \hline \hline

\multirow{4}{*}{062-dice}
& FSG               & 0.026                      & 8.09$\text{e-03}$                      & 5.62$\text{e-04}$                      & 16.5\%                      \\
& FSG+Aff           & 0.025                      & 8.65$\text{e-03}$                      & 5.74$\text{e-04}$                      & 14.8\%                      \\
& FSG+Aff-FinIter   & \underline{\textbf{0.024}} & \underline{\textbf{8.76$\text{e-03}$}} & \underline{\textbf{5.83$\text{e-04}$}} & 14.2\%                      \\
& FSG+Aff+LinkPen   & 0.029                      & 8.59$\text{e-03}$                      & 5.58$\text{e-04}$                      & \underline{\textbf{16.7\%}} \\ \hline \hline

\end{tabular}
}
\end{table}

For all the valid YCB objects, the collected grasp pose demonstrations and the augmented grasp poses with finger iteration and link penetration checking are illustrated in Fig.\ref{fig:aug1} and Fig.\ref{fig:aug2}.

\begin{figure}
  \centering
  \includegraphics[width=\columnwidth]{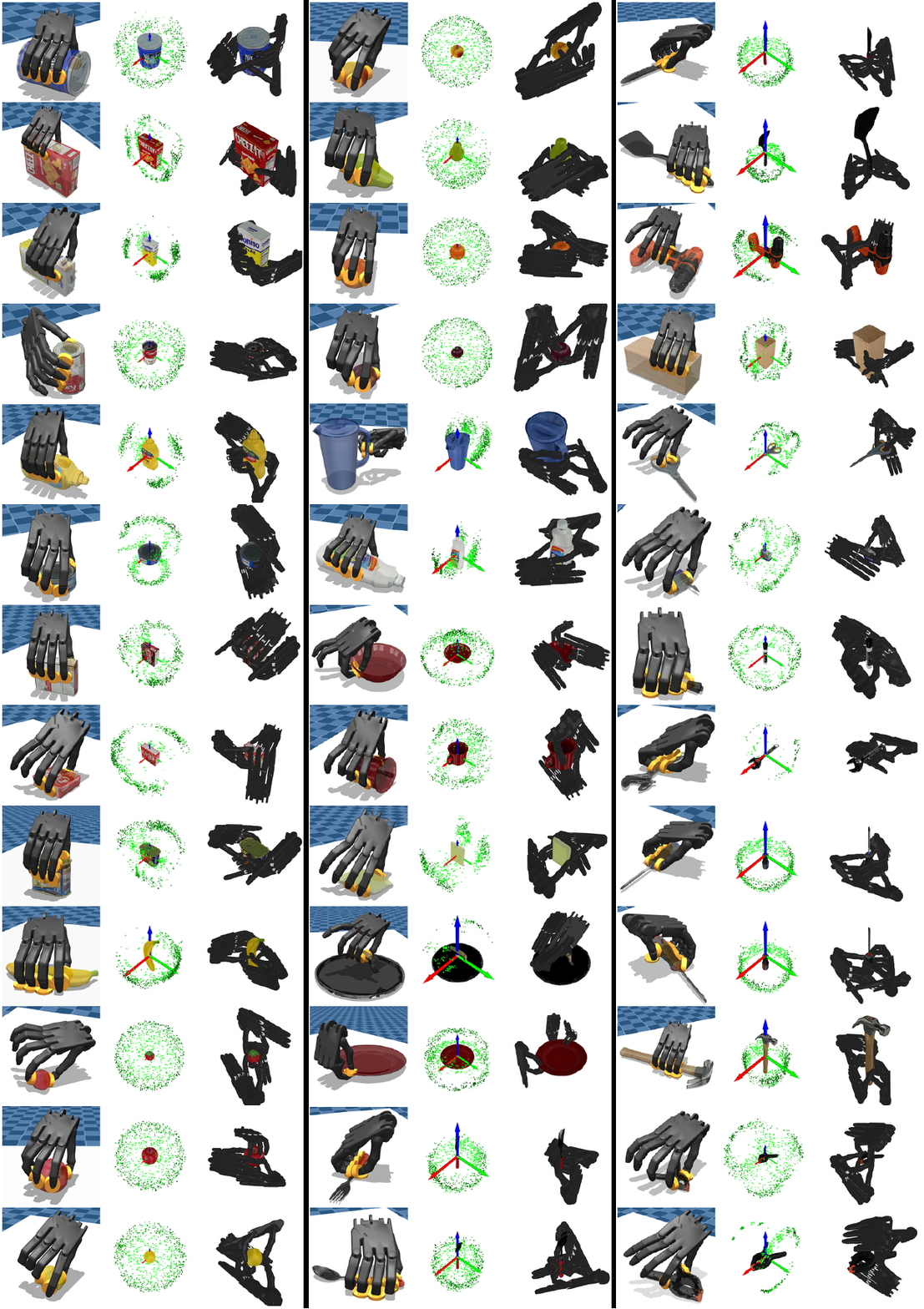}
  \caption{Collected grasp pose demonstrations and augmented datasets (Part1). For each object, the left panel shows the grasp pose demonstration, with yellow cylinders denoting the contact points, the middle panel shows the point cloud of all palm positions from the augmented datasets, and the right panel shows three example grasp poses.
  }
  \label{fig:aug1}
\end{figure}

\begin{figure}
  \centering
  \includegraphics[width=\columnwidth]{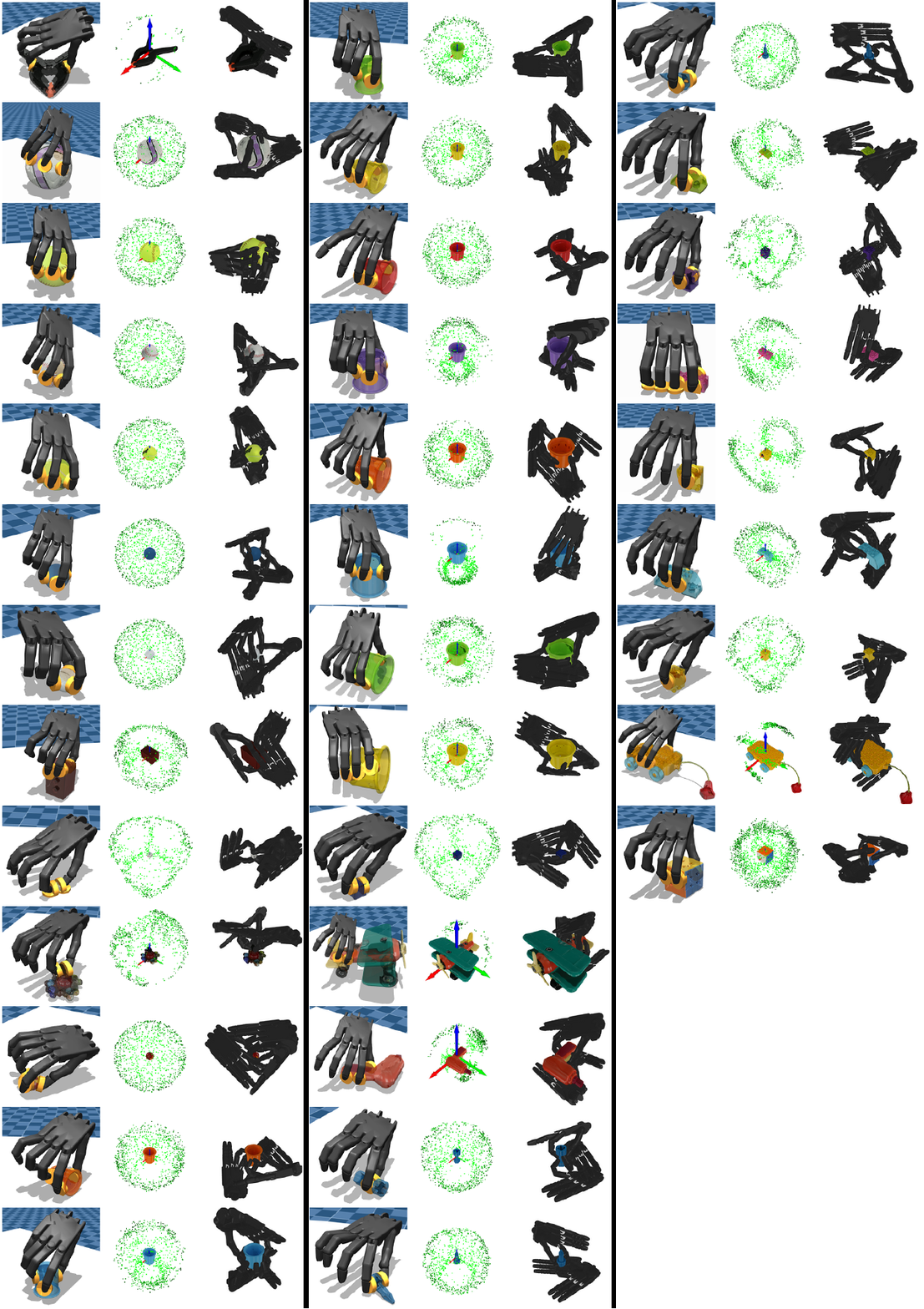}
  \caption{Collected grasp pose demonstrations and augmented datasets (Part2). For each object, the left panel shows the grasp pose demonstration, with yellow cylinders denoting the contact points, the middle panel shows the point cloud of all palm positions from the augmented datasets, and the right panel shows three example grasp poses.
  }
  \label{fig:aug2}
\end{figure}

\subsection{Hand-Arm Grasp Data Generation}

In Sec.\ref{sec:gdg} and Sec.\ref{sec:gda}, we focused merely on the interaction between the robotic hand and the object. However, in practical applications, it is essential to also consider the arm's reachable workspace and potential collisions between the robot and the environment. In this section, we integrate our proposed method with robotic arm trajectory planning to generate grasp data that include arm trajectories.

Starting with the hand-only grasp data, objects are randomly placed on a table and allowed to stabilize. The grasp data is then transformed to align with the corresponding object pose. Grasp poses that result in collisions between the hand and the table are filtered out, and IK and trajectory planning are applied to the remaining poses. The IK solutions with grasp poses are simulated in MuJoCo to evaluate the configurations of the robotic arm, hand, and object. The validity of the generated poses is assessed based on changes in the object pose after grasping. Examples of the generated poses are shown in Fig.\ref{fig:hand_arm}. These data can be used for data-driven grasp pose prediction, as target states for dynamic grasp motion planning and as initial states for manipulation learning.

\begin{figure}
  \centering
  \includegraphics[width=\columnwidth]{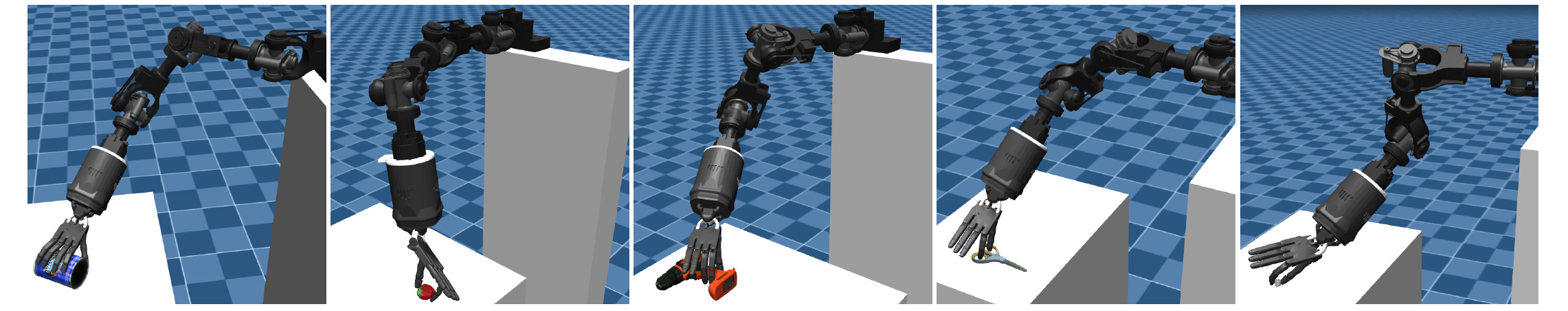}
  \caption{Grasp poses for objects on a table using a hand-arm system.}
  \label{fig:hand_arm}
\end{figure}

}
{
}

\section{Conclusion} \label{sec:conclusion}

\switchlanguage%
{

In this work, we introduced a pipeline for collecting grasp demonstrations through teleoperation, using AutoWS to generate fingertip workspace clouds based on the hand model and the demonstrated grasp pose, and employing FSG to augment the collected grasp data. By representing each fingertip workspace independently as a point cloud, we can easily configure the number of fingers used for grasping. When incorporating the demonstrated grasp pose to generate workspace clouds, the points in the generated cloud concentrate within the known valid range, reducing the solution space to a more compact valid region. This improves the speed and rate of valid grasp generation, while also enabling the handling of small or thin objects, as shown in comparisons with other methods. The FSG module directly operates on the workspace clouds, making it applicable to robotic hands with arbitrary mechanical structures. With its simple configuration and efficiency in grasp generation and augmentation, we hope our method will assist in robot grasp data expansion and support the training of data-driven methods.

The current approach still has some limitations. For example, it requires the VR equipment to collect grasping data, which makes the process relatively complex. Simplifying the workflow by directly retargeting hand poses from images or videos to robotic hands and estimating contact points would be a significant improvement. While our method focuses on generating data for precision grasps, future work will explore how to generate corresponding postures for power grasps or transition between different types of grasps. Furthermore, efficiently reconstructing the complete point cloud of an object in real-world environments will extend the applicability of our approach to real robot grasping tasks.

}
{
}

{
\bibliographystyle{junsrt}
\bibliography{main}

\begin{thebibliography}{10}

\bibitem{anti1}
Jeffrey Mahler, Jacky Liang, Sherdil Niyaz, Michael Laskey, Richard Doan, Xinyu
  Liu, Juan~Aparicio Ojea, and Ken Goldberg.
\newblock Dex-net 2.0: Deep learning to plan robust grasps with synthetic point
  clouds and analytic grasp metrics.
\newblock In {\em Robotics: Science and Systems (RSS)}, 2017.

\bibitem{fin21}
Yunzhi Lin, Chao Tang, Fu-Jen Chu, and Patricio~A. Vela.
\newblock Using synthetic data and deep networks to recognize primitive shapes
  for object grasping.
\newblock In {\em 2020 IEEE International Conference on Robotics and Automation
  (ICRA)}, pp. 10494--10501, 2020.

\bibitem{fin22}
Yunho Choi, Hogun Kee, Kyungjae Lee, JaeGoo Choy, Junhong Min, Sohee Lee, and
  Songhwai Oh.
\newblock Hierarchical 6-dof grasping with approaching direction selection.
\newblock In {\em 2020 IEEE International Conference on Robotics and Automation
  (ICRA)}, pp. 1553--1559, 2020.

\bibitem{fin23}
Martin Sundermeyer, Arsalan Mousavian, Rudolph Triebel, and Dieter Fox.
\newblock Contact-graspnet: Efficient 6-dof grasp generation in cluttered
  scenes.
\newblock In {\em 2021 IEEE International Conference on Robotics and Automation
  (ICRA)}, 2021.

\bibitem{approach1}
Philipp Schmidt, Nikolaus Vahrenkamp, Mirko W^^c3^^a4chter, and Tamim Asfour.
\newblock Grasping of unknown objects using deep convolutional neural networks
  based on depth images.
\newblock In {\em 2018 IEEE International Conference on Robotics and Automation
  (ICRA)}, pp. 6831--6838, 2018.

\bibitem{exhau3}
Kelin Li, Nicholas Baron, Xian Zhang, and Nicolas Rojas.
\newblock Efficientgrasp: A unified data-efficient learning to grasp method for
  multi-fingered robot hands.
\newblock {\em IEEE Robotics and Automation Letters}, Vol.~7, No.~4, pp.
  8619--8626, 2022.

\bibitem{primitive}
Qingkai Lu, Mark Van~der Merwe, Balakumar Sundaralingam, and Tucker Hermans.
\newblock Multifingered grasp planning via inference in deep neural networks:
  Outperforming sampling by learning differentiable models.
\newblock {\em IEEE Robotics \& Automation Magazine}, Vol.~27, No.~2, pp.
  55--65, 2020.

\bibitem{fast}
Dylan Turpin, Tao Zhong, Shutong Zhang, Guanglei Zhu, Eric Heiden, Miles
  Macklin, Stavros Tsogkas, Sven Dickinson, and Animesh Garg.
\newblock Fast-grasp'd: Dexterous multi-finger grasp generation through
  differentiable simulation.
\newblock In {\em 2023 IEEE International Conference on Robotics and Automation
  (ICRA)}, 2023.

\bibitem{dex}
Ruicheng Wang, Jialiang Zhang, Jiayi Chen, Yinzhen Xu, Puhao Li, Tengyu Liu,
  and He~Wang.
\newblock Dexgraspnet: A large-scale robotic dexterous grasp dataset for
  general objects based on simulation.
\newblock In {\em 2023 IEEE International Conference on Robotics and Automation
  (ICRA)}, pp. 11359--11366, 2023.

\bibitem{billion}
Clemens Eppner, Arsalan Mousavian, and Dieter Fox.
\newblock A billion ways to grasps - an evaluation of grasp sampling schemes on
  a dense, physics-based grasp data set.
\newblock In {\em Proceedings of the International Symposium on Robotics
  Research ({ISRR})}, Hanoi, Vietnam, 2019.

\bibitem{pds}
Wu~Albert, Guo Michelle, and Liu C.~Karen.
\newblock Learning diverse and physically feasible dexterous grasps with
  generative model and bilevel optimization.
\newblock In {\em 6th Conference on Robot Learning (CoRL 2022)}, 2022.

\bibitem{tele}
Xuxin Cheng, Jialong Li, Shiqi Yang, Ge~Yang, and Xiaolong Wang.
\newblock Open-television: Teleoperation with immersive active visual feedback.
\newblock {\em arXiv preprint arXiv:2407.01512}, 2024.

\bibitem{aug}
Shorten Connor and Khoshgoftaar Taghi, M.
\newblock A survey on image data augmentation for deep learning.
\newblock {\em Journal of Big Data}, 2019.

\bibitem{antipodal}
G.~Smith, E.~Lee, K.~Goldberg, K.~Bohringer, and J.~Craig.
\newblock Computing parallel-jaw grips.
\newblock In {\em Proceedings 1999 IEEE International Conference on Robotics
  and Automation (Cat. No.99CH36288C)}, Vol.~3, pp. 1897--1903 vol.3, 1999.

\bibitem{anti2}
Chaozheng Wu, Jian Chen, Qiaoyu Cao, Jianchi Zhang, Yunxin Tai, Lin Sun, and
  Kui Jia.
\newblock Grasp proposal networks: an end-to-end solution for visual learning
  of robotic grasps.
\newblock In {\em Proceedings of the 34th International Conference on Neural
  Information Processing Systems}, NIPS'20, 2020.

\bibitem{anti3}
Clemens Eppner, Arsalan Mousavian, and Dieter Fox.
\newblock Acronym: A large-scale grasp dataset based on simulation.
\newblock In {\em 2021 IEEE International Conference on Robotics and Automation
  (ICRA)}, pp. 6222--6227, 2021.

\bibitem{anti4}
Wei Wei, Yongkang Luo, Fuyu Li, Guangyun Xu, Jun Zhong, Wanyi Li, and Peng
  Wang.
\newblock Gpr: Grasp pose refinement network for cluttered scenes.
\newblock In {\em 2021 IEEE International Conference on Robotics and Automation
  (ICRA)}, pp. 4295--4302. IEEE Press, 2021.

\bibitem{exhau1}
Hongzhuo Liang, Xiaojian Ma, Shuang Li, Michael G^^c3^^b6rner, Song Tang, Bin
  Fang, Fuchun Sun, and Jianwei Zhang.
\newblock Pointnetgpd: Detecting grasp configurations from point sets.
\newblock In {\em 2019 International Conference on Robotics and Automation
  (ICRA)}, pp. 3629--3635, 2019.

\bibitem{exhau2}
Lin Shao, Fabio Ferreira, Mikael Jorda, Varun Nambiar, Jianlan Luo, Eugen
  Solowjow, Juan~Aparicio Ojea, Oussama Khatib, and Jeannette Bohg.
\newblock Unigrasp: Learning a unified model to grasp with multifingered
  robotic hands.
\newblock {\em IEEE Robotics and Automation Letters}, Vol.~5, No.~2, pp.
  2286--2293, 2020.

\bibitem{approach}
N.~Vahrenkamp, M.~Kr{\"o}hnert, S.~Ulbrich, T.~Asfour, G.~Metta, R.~Dillmann,
  and G.~Sandini.
\newblock {\em Simox: A Robotics Toolbox for Simulation, Motion and Grasp
  Planning}, pp. 585--594.
\newblock Springer Berlin Heidelberg, Berlin, Heidelberg, 2013.

\bibitem{approach2}
Matt Corsaro, Stefanie Tellex, and George Konidaris.
\newblock Learning to detect multi-modal grasps for dexterous grasping in dense
  clutter.
\newblock In {\em 2021 IEEE/RSJ International Conference on Intelligent Robots
  and Systems (IROS)}, pp. 4647--4653. IEEE Press, 2021.

\bibitem{approach3}
Wei Wei, Daheng Li, Peng Wang, Yiming Li, Wanyi Li, Yongkang Luo, and Jun
  Zhong.
\newblock Dvgg: Deep variational grasp generation for dextrous manipulation.
\newblock {\em IEEE RAL}, Vol.~7, No.~2, pp. 1659--1666, 2022.

\bibitem{approach4}
Feng Q., Lema D., Malmir M., Li~H., Feng J., Chen Z., and Knoll A.
\newblock Dexgangrasp: Dexterous generative adversarial grasping synthesis for
  task-oriented manipulation.
\newblock In {\em 2024 IEEE-RAS 23rd International Conference on Humanoid
  Robots (Humanoids)}. IEEE Press, 2024.

\bibitem{qd}
Huber J., H^^c3^^a9l^^c3^^a9non F., Kappel M., Chelly E., Khoramshahi M., Amar
  F., Ben, and Doncieux S.
\newblock Speeding up 6-dof grasp sampling with quality-diversity.
\newblock In {\em 2024 IEEE/RSJ International Conference on Intelligent Robots
  and Systems (IROS)}. IEEE Press, 2024.

\bibitem{eigen}
Matei Ciocarlie, Corey Goldfeder, and Peter Allen.
\newblock Dimensionality reduction for hand-independent dexterous robotic
  grasping.
\newblock In {\em 2007 IEEE/RSJ International Conference on Intelligent Robots
  and Systems}, pp. 3270--3275, 2007.

\bibitem{eigen1}
Jens Lundell, Francesco Verdoja, and Ville Kyrki.
\newblock Robust grasp planning over uncertain shape completions.
\newblock In {\em 2019 IEEE/RSJ International Conference on Intelligent Robots
  and Systems (IROS)}, pp. 1526--1532, 2019.

\bibitem{eigen2}
Liu Min, Pan Zherong, Xu~Kai, Ganguly Kanishka, and Manocha Dinesh.
\newblock Deep differentiable grasp planner for high-dof grippers.
\newblock In {\em Robotics: Science and Systems (RSS)}, 2020.

\bibitem{eigen3}
Mia Kokic, Danica Kragic, and Jeannette Bohg.
\newblock Learning task-oriented grasping from human activity datasets.
\newblock {\em IEEE Robotics and Automation Letters}, Vol.~5, No.~2, pp.
  3352--3359, 2020.

\bibitem{eigen4}
Jens Lundell, Francesco Verdoja, and Ville Kyrki.
\newblock {{DDGC}}: {{Generative Deep Dexterous Grasping}} in {{Clutter}}.
\newblock {\em IEEE Robotics and Automation Letters}, Vol.~6, No.~4, pp.
  6899--6906, October 2021.

\bibitem{dex1}
Jiaxin Lu, Hao Kang, Haoxiang Li, Bo~Liu, Yiding Yang, Qi-Xing Huang, and Gang
  Hua.
\newblock Ugg: Unified generative grasping.
\newblock In {\em European Conference on Computer Vision}, 2023.

\bibitem{dex2}
Guo-Hao Xu, Yi-Lin Wei, Dian Zheng, Xiao-Ming Wu, and Wei-Shi Zheng.
\newblock Dexterous grasp transformer.
\newblock In {\em Proceedings of the IEEE/CVF Conference on Computer Vision and
  Pattern Recognition}, 2024.

\bibitem{contact}
Puhao Li, Tengyu Liu, Yuyang Li, Yiran Geng, Yixin Zhu, Yaodong Yang, and
  Siyuan Huang.
\newblock Gendexgrasp: Generalizable dexterous grasping.
\newblock {\em 2023 IEEE International Conference on Robotics and Automation
  (ICRA)}, pp. 8068--8074, 2022.

\bibitem{contact1}
Yinzhen Xu, Weikang Wan, Jialiang Zhang, Haoran Liu, Zikang Shan, Hao Shen,
  Ruicheng Wang, Haoran Geng, Yijia Weng, Jiayi Chen, Tengyu Liu, Li~Yi, and
  He~Wang.
\newblock Unidexgrasp: Universal robotic dexterous grasping via learning
  diverse proposal generation and goal-conditioned policy.
\newblock In {\em 2023 IEEE/CVF Conference on Computer Vision and Pattern
  Recognition (CVPR)}, pp. 4737--4746, 2023.

\bibitem{contact2}
Fuqiang Zhao, Dzmitry Tsetserukou, and Qian Liu.
\newblock Graingrasp: Dexterous grasp generation with fine-grained contact
  guidance.
\newblock In {\em 2024 IEEE International Conference on Robotics and Automation
  (ICRA)}, 2024.

\bibitem{poisson}
Cem Yuksel.
\newblock {Sample Elimination for Generating Poisson Disk Sample Sets}.
\newblock {\em Computer Graphics Forum}, 2015.

\bibitem{hidden}
Sagi Katz, Ayellet Tal, and Ronen Basri.
\newblock Direct visibility of point sets.
\newblock {\em ACM Trans. Graph.}, Vol.~26, No.~3, pp. 24--36, jul 2007.

\bibitem{curv}
Crane~He Chen.
\newblock Estimating discrete total curvature with per triangle normal
  variation.
\newblock In {\em ACM SIGGRAPH 2023 Talks}, SIGGRAPH '23, New York, NY, USA,
  2023. Association for Computing Machinery.

\bibitem{gws}
C.~Ferrari and J.~Canny.
\newblock Planning optimal grasps.
\newblock In {\em Proceedings 1992 IEEE International Conference on Robotics
  and Automation}, pp. 2290--2295 vol.3, 1992.

\bibitem{ycb}
Berk Calli, Arjun Singh, Aaron Walsman, Siddhartha Srinivasa, Pieter Abbeel,
  and Aaron~M. Dollar.
\newblock The ycb object and model set: Towards common benchmarks for
  manipulation research.
\newblock In {\em 2015 International Conference on Advanced Robotics (ICAR)},
  pp. 510--517, 2015.

\bibitem{shadow}
{ShadowRobot} {S}hadow{R}obot {D}exterous {H}and.
\newblock \url{www.shadowrobot.com/dexterous-hand-series/}.

\bibitem{mujoco}
Emanuel Todorov, Tom Erez, and Yuval Tassa.
\newblock Mujoco: A physics engine for model-based control.
\newblock In {\em 2012 IEEE/RSJ International Conference on Intelligent Robots
  and Systems}, pp. 5026--5033. IEEE, 2012.

\bibitem{graspit}
A.T. Miller and P.K. Allen.
\newblock Graspit! a versatile simulator for robotic grasping.
\newblock {\em IEEE Robotics \& Automation Magazine}, Vol.~11, No.~4, pp.
  110--122, 2004.

\bibitem{robotiq}
{ROBOTIQ} {R}obotiq 2{F}-85/140.
\newblock \url{https://robotiq.com/products/2f85-140-adaptive-robot-gripper}.

\bibitem{barrett}
{BarrettHand} {B}arrett {H}and.
\newblock \url{https://barrett.com/barrett-hand}.

\bibitem{allegro}
{WONIK ROBOTICS} {A}llegro {H}and.
\newblock \url{https://www.wonikrobotics.com/robot-hand}.

\end{thebibliography}
}

\end{document}